\documentclass[10pt,sigconf,letterpaper]{acmart}
\usepackage[utf8]{inputenc}
\usepackage{amsmath}
\usepackage{graphicx}
\usepackage{enumitem}
\usepackage{xcolor}
\usepackage{hyperref}
\usepackage{booktabs}
\usepackage{tabularx}
\usepackage{tikz,graphics,color,float,epsf,caption,subcaption}
\usepackage{balance}

\graphicspath{{figures/}}

\setcopyright{none}
\renewcommand\footnotetextcopyrightpermission[1]{} 
\begin{document}
\acmYear{}
\copyrightyear{}
\setcopyright{acmcopyright}
\acmConference{Generative Models}{GANs and VAEs}{Network Anomaly Detection}
\acmPrice{}
\acmDOI{}
\acmISBN{}

\title[Generative Models for Network Anomaly Detection]{On the Usage of Generative Models for Network Anomaly Detection in Multivariate Time-Series}

\author{Gast\'on Garc\'ia Gonz\'alez}
\affiliation{
  \institution{IE--FING, UDELAR, Uruguay\\AIT Austrian Institute of Technology}
}
\email{gastong@fing.edu.uy}

\author{Pedro Casas}
\affiliation{
  \institution{AIT Austrian Institute of Technology}
}
\email{pedro.casas@ait.ac.at}

\author{Alicia Fern\'andez}
\affiliation{
  \institution{IE--FING, UDELAR, Uruguay}
}
\email{alicia@fing.edu.uy}

\author{Gabriel G\'omez}
\affiliation{
  \institution{IE--FING, UDELAR, Uruguay}
}
\email{ggomez@fing.edu.uy}

\renewcommand{\shortauthors}{G. Garc\'ia Gonz\'alez, P. Casas, A. Fern\'andez, G. G\'omez}

\begin{abstract}
Despite the many attempts and approaches for anomaly detection explored over the years, the automatic detection of rare events in data communication networks remains a complex problem. In this paper we introduce \emph{Net-GAN}, a novel approach to network anomaly detection in time-series, using recurrent neural networks (RNNs) and generative adversarial networks (GAN). Different from the state of the art, which traditionally focuses on univariate measurements, Net-GAN detects anomalies in multivariate time-series, exploiting temporal dependencies through RNNs. Net-GAN discovers the underlying distribution of the baseline, multivariate data, without making any assumptions on its nature, offering a powerful approach to detect anomalies in complex, difficult to model network monitoring data. We further exploit the concepts behind generative models to conceive Net-VAE, a complementary approach to Net-GAN for network anomaly detection, based on variational auto-encoders (VAE). We evaluate Net-GAN and Net-VAE in different monitoring scenarios, including anomaly detection in IoT sensor data, and intrusion detection in network measurements. Generative models represent a promising approach for network anomaly detection, especially when considering the complexity and ever-growing number of time-series to monitor in operational networks.
\end{abstract}

\maketitle

\section{Introduction}

Network monitoring data generally consists of hundreds or thousands of counters periodically collected in the form of time-series, resulting in a complex-to-analyze multivariate time-series process (MTS). In particular, detecting anomalies in such multivariate, temporal data is challenging. Without loss of generality, we refer to the MTS as a set of $n$, non-iid time series sampled at the same rate, referred to as $x_t = \{x_t(1), x_t(2),\dots, x_t(n)\} \in {\rm I\!R}^n$. Current approaches to anomaly detection tackle this challenge by either focusing on univariate time-series analysis -- running an independent detector for each time-series $x_t(i)$, or by considering multi-dimensional input data $x \in {\rm I\!R}^n$ at each time $t$, neglecting the temporal aspects of the MTS. To improve the state of affairs we propose Net-GAN, a novel unsupervised approach to anomaly detection in MTS data, based on Recurrent Neural Networks (RNNs), trained through a Generative Adversarial Networks framework (GAN) \cite{goodfellow2014}.

The usage of generative models for semi-supervised anomaly detection helps to solve two major problems faced in this specific field: the high imbalance between normal operation and anomaly instances, as well as the lack of labeled instances for learning and validation purposes. Generative models such as Variational Auto-Encoders (VAEs) or Generative Adversarial Networks (GANs) are powerful approaches to learn the underlying distributions of data samples, in a purely data-driven, model-agnostic manner. Such models can be used in the practice to construct better baselines (i.e., profiles for normal operation) for the anomaly detection task, improving the identification of instances which deviate from this baseline. Most of previous work in this direction treats data as temporally independent samples, neglecting the information provided by causality and temporal correlation.

To capture the temporal correlations characterizing an MTS, we adapt the original GAN model proposed in \cite{goodfellow2014}, replacing the multilayer perceptrons by recursive, LSTM networks for both generator and discriminator models. The input data is therefore sequences of multi-dimensional measurements, of length $T$: $\lbrace x_{t-T}, ..., x_t\rbrace$. In a similar direction, we also explore the performance of other powerful generative models for anomaly detection, using in particular VAEs. Variational auto-encoders are a generative version of classical auto-encoders, but different from GANs, they make strong assumptions on the generative distribution they try to estimate. We refer to this flavor of Net-GAN as Net-VAE.  

The reminder of the paper is organized as follows: Section \ref{SII} briefly overviews the related work; in Section \ref{SIII}, we describe the Net-GAN and Net-VAE approaches; Section \ref{SIV} reports the preliminary evaluation results obtained with Net-GAN/Net-VAE models in the detection of anomalies in different datasets. Finally, Section \ref{SV} concludes the paper.

\begin{figure}[t!]
  \centering
  \includegraphics[width=0.9\linewidth]{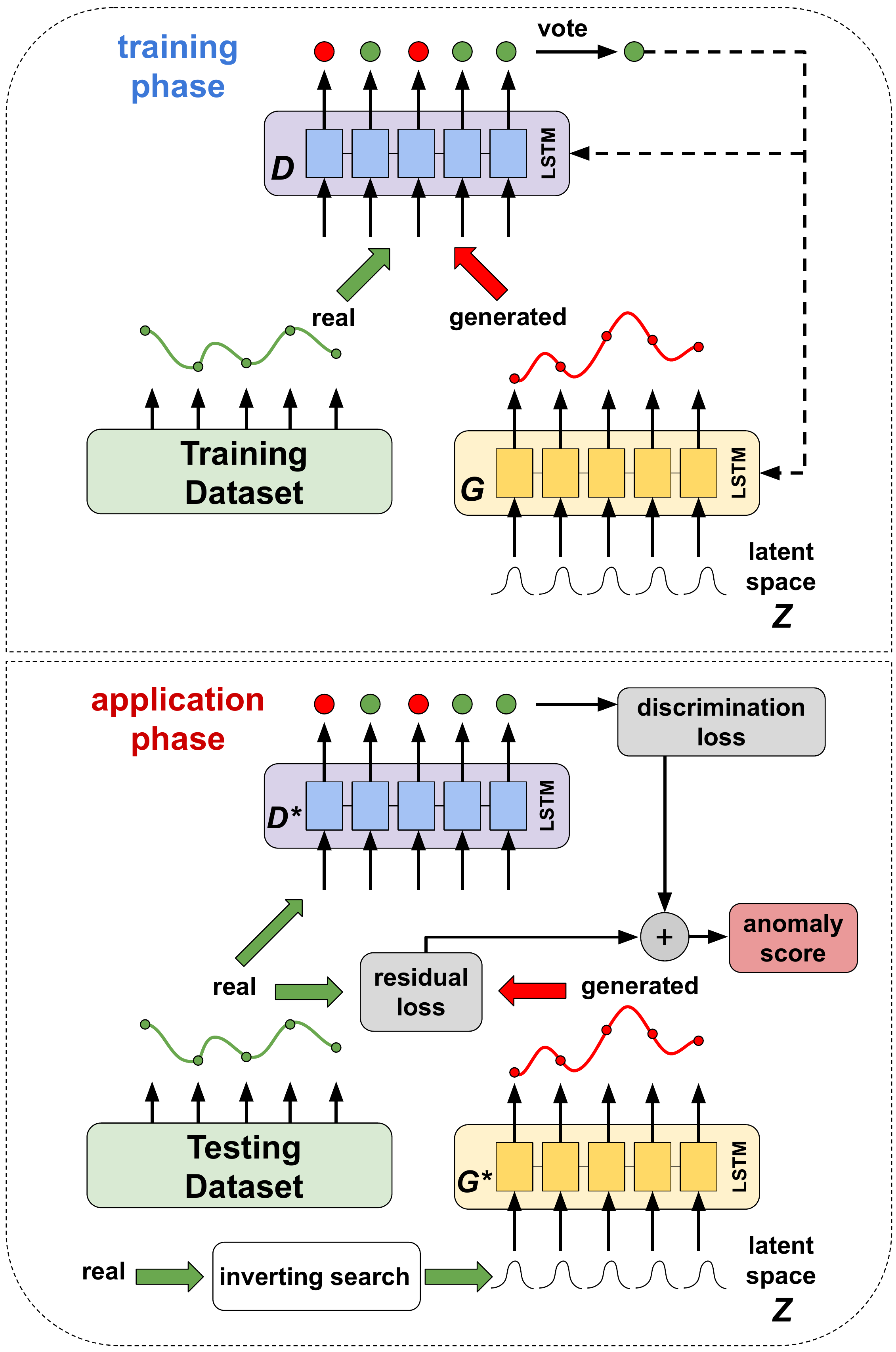}
  \caption{Net-GAN architecture and its application.}
  \label{Net-GAN}
\end{figure}

\section{Related Work}\label{SII}

Generally speaking, operational network monitoring systems rely on rules and fingerprints to detect anomalous behaviors. There are nevertheless multiple extensive surveys on general domain anomaly detection techniques~\cite{chandola2009anomaly} as well as on network anomaly detection~\cite{ahmed2016survey,zhang2009survey}, including machine learning-based approaches. There is a particularly extensive literature in the application of learning-based approaches for automatic traffic monitoring and analysis~\cite{Boutaba2018}, including detection. Their main limitation as compared to our work is their (generally) supervised nature, which requires ground-truth data for learning. There is also a vast literature on clustering--based approaches for unsupervised network anomaly detection and analysis, mostly targeting the security domain \cite{casas2012unsupervised, dromard2016online, goldstein2016comparative}.

When it comes to the application of generative models for anomaly detection, there are recent papers on GANs for time-series synthesizing and anomaly detection \cite{di2019survey, esteban2017real, li2018anomaly}. Examples of GANs for anomaly detection, as well as VAEs for anomaly detection, are presented in \cite{zavrak2020anomaly} and \cite{zenati2018efficient}, respectively. A particularly interesting model for anomaly detection using GANs is BiGAN \cite{bigan2016}, which extends the original GAN architecture by adding the learning of the inverse mapping which maps the data back to the latent representation. 

\begin{figure}[t!]
  \centering
  \includegraphics[width=0.9\linewidth]{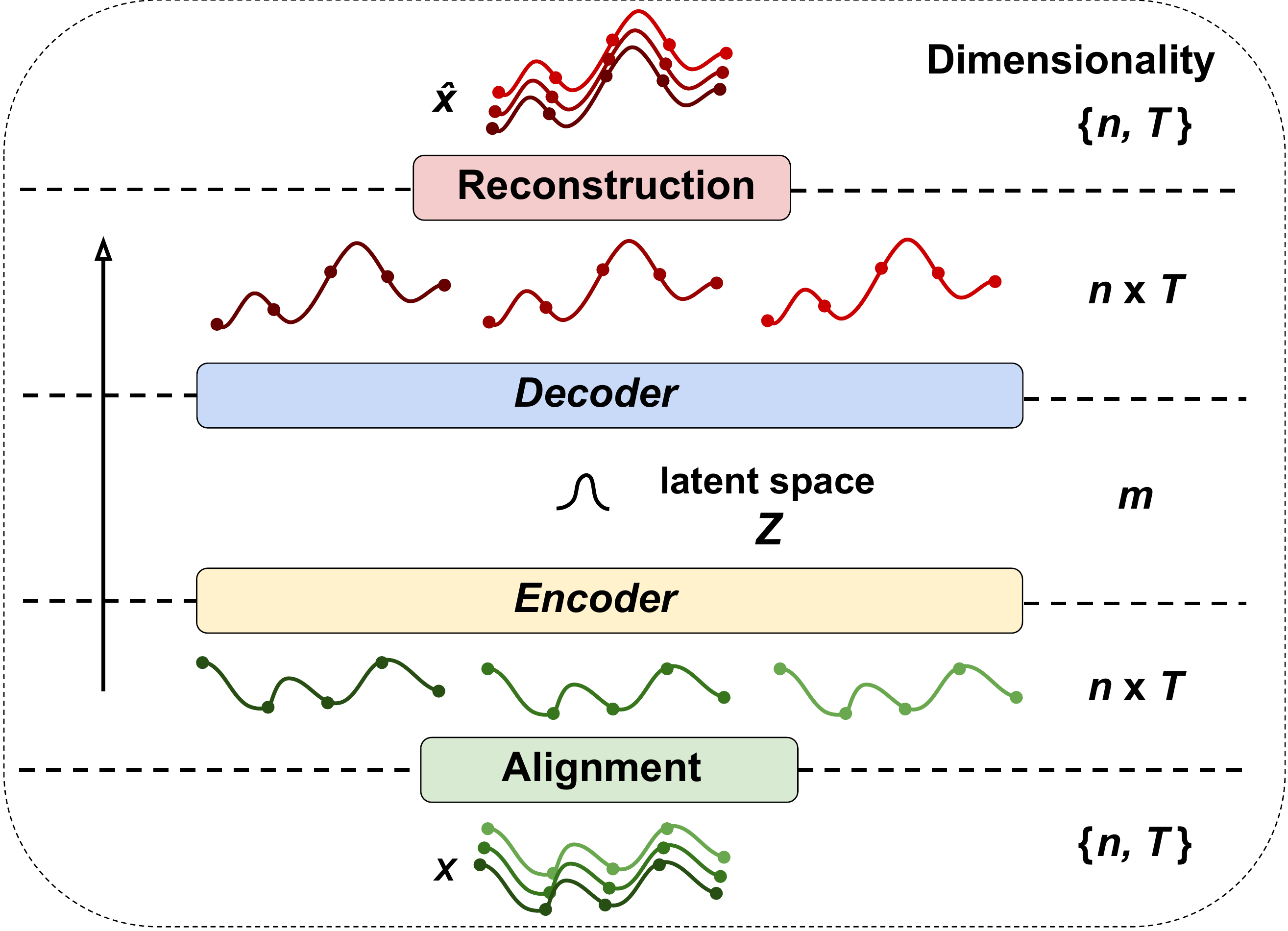}
  \caption{Net-VAE architecture.}
  \label{Net-VAE}
\end{figure}

\section{The Net-GAN/VAE Approach}\label{SIII}

GANs are a framework for the estimation of generative models via an adversarial process in which two models, a discriminator $D$ and a generator $G$, are trained simultaneously, in an adversarial manner. The generator $G$ aim is to capture the -- unknown and potentially complex, data distribution, while the discriminator $D$ estimates the probability that
a sample came from the training data rather than $G$. To learn a generative distribution $p_g$ over the learning data $x$, the generator builds a mapping from a prior noise distribution $p_z$ to a data space as $G(z)$. The discriminator outputs a single scalar $D(x)$ representing the probability that input $x$ came from real data rather than from $p_g$.

Fig.~\ref{Net-GAN} depicts the Net-GAN architecture and both the model training and anomaly detection procedures. In the \textbf{\emph{training phase}} (top), the generator $G$ draws synthetic sample sequences $G(z)$ from Gaussian noise -- the latent space $Z$, with the objective of deceiving the discriminator $D$, which in turn learns to determine whether training samples are real or derived from the generative distribution. The classification result proposed by $D$ is additionally fed back to $G$, serving as a reinforcement loop to guide the generation process. As both $G$ and $D$ compete to achieve their adversarial tasks, synthetic samples become more and more ``realistic'', and the discriminator becomes robust to noise, improving the detection of non-conforming (i.e., out of the baseline) samples. 

\begin{figure}[t!]
  \centering
  \includegraphics[width=\linewidth]{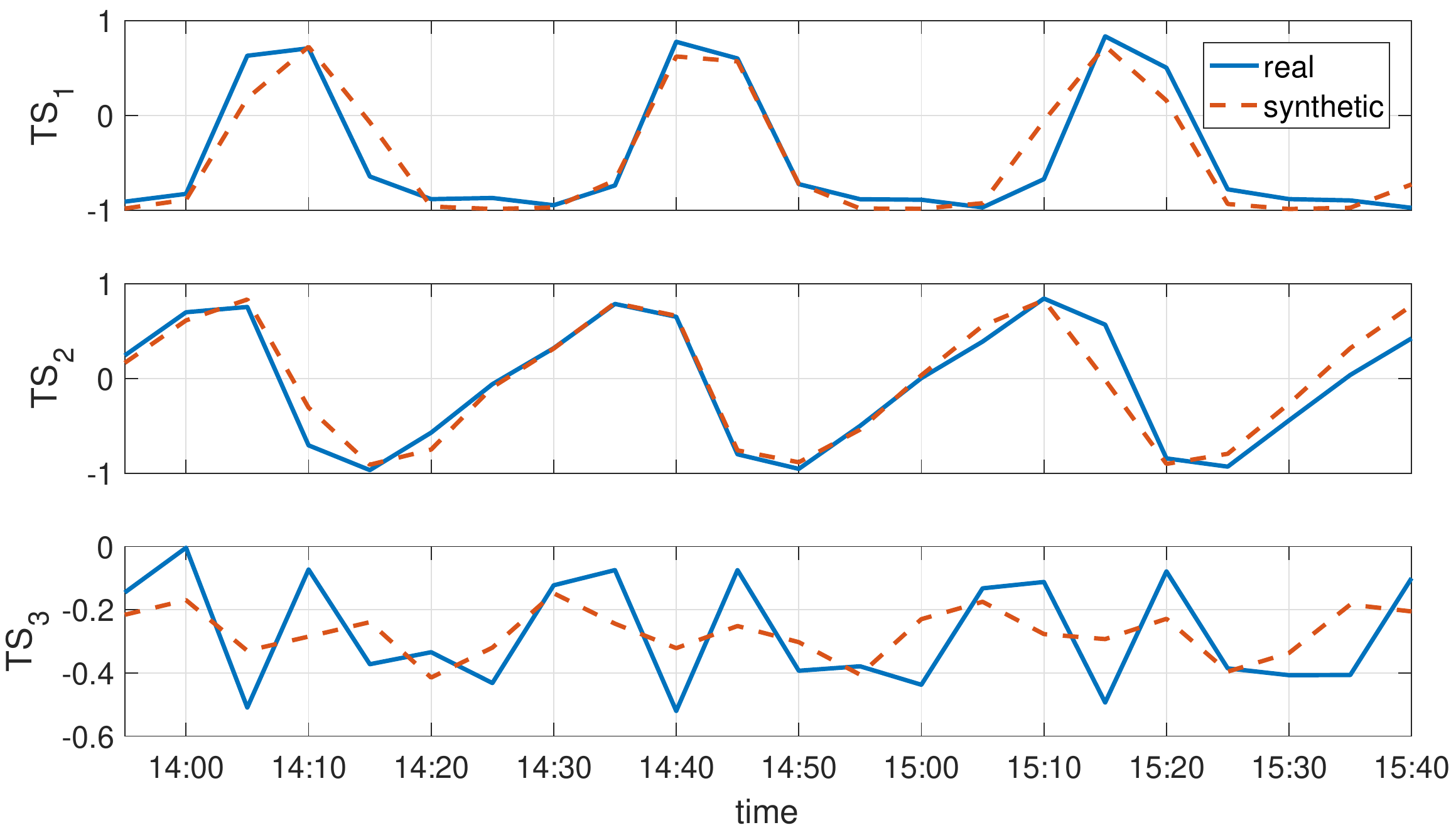}
  \caption{Net-GAN: synthetically generated time-series.}
  \label{synthetic_ts}
\end{figure}

In the \textbf{\emph{application phase}} (down), the trained discriminator $D^*$ acts naturally as an anomaly detector, detecting deviations from the baseline, through a \emph{discrimination loss} function. The trained generator $G^*$ is also used to improve detection performance, serving as baseline generation; by doing an inverse search in the latent space -- for example, constructing an inverse model for the generator \cite{bigan2016,zenati2018efficient}, we find the sample $z \in Z$ which generates the closest sample $\hat{x}$ to the tested one $x$, producing a \emph{residual loss}. This step also be approximated by randomly sampling the latent space, and keeping the sample $\hat{x}$ which better approximates $x$. Both the discrimination and the residual loss functions can be combined into an \emph{anomaly score}, which is compared to a calibrated threshold to take the final decision. 

As we mentioned before, one of the salient features of Net-GAN is that we use LSTM networks for both $G$ and $D$, instead of the traditionally used multilayer perceptrons or convolutional neural networks. Being recursive by conception, LSTMs can capture temporal dependencies in the data, which feed-forward networks fail to do. This is paramount when it comes to time-series analysis. In Section \ref{SIV} we test separately the detection performance provided by the trained discriminator $D^*$ and the trained generator $G^*$, using random sampling as reverse-search technique.

\begin{figure}[t!]
\renewcommand{\arraystretch}{0.7}
\centering
$\begin{array}{c}
\includegraphics[width=0.95\linewidth]{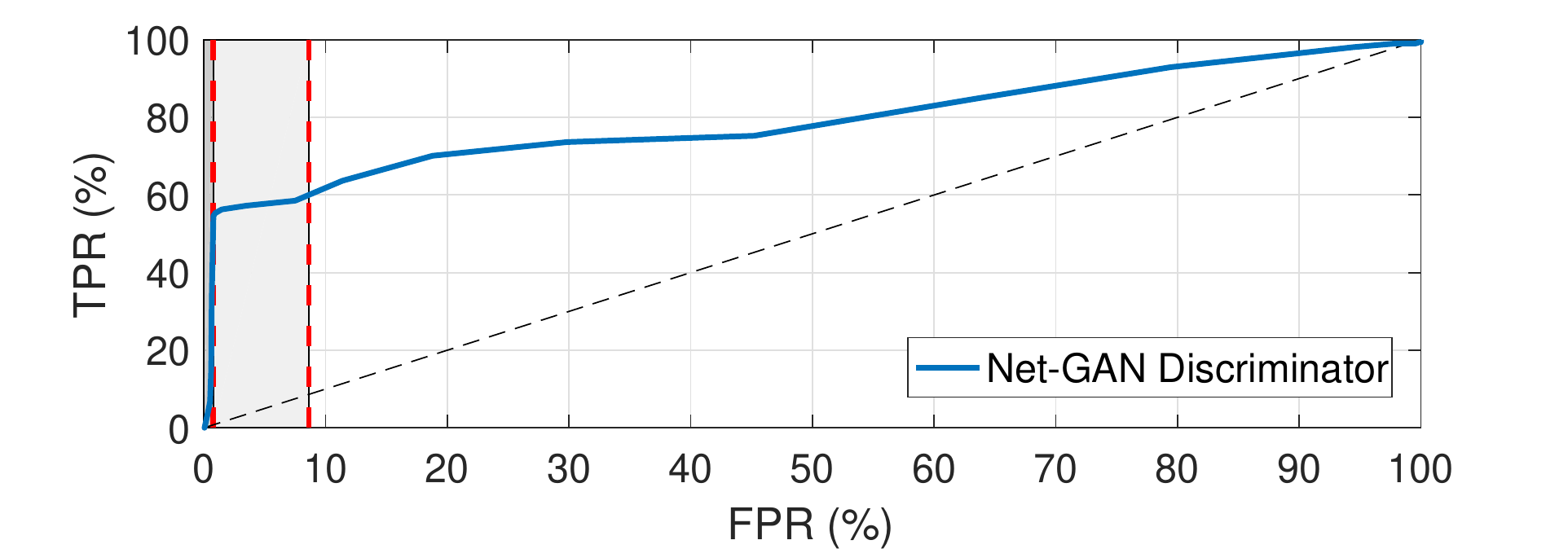}\\\vspace{-0.2mm}
\text{\scriptsize{(a) Detection performance with Net-GAN-D.}}\\
\includegraphics[width=0.95\linewidth]{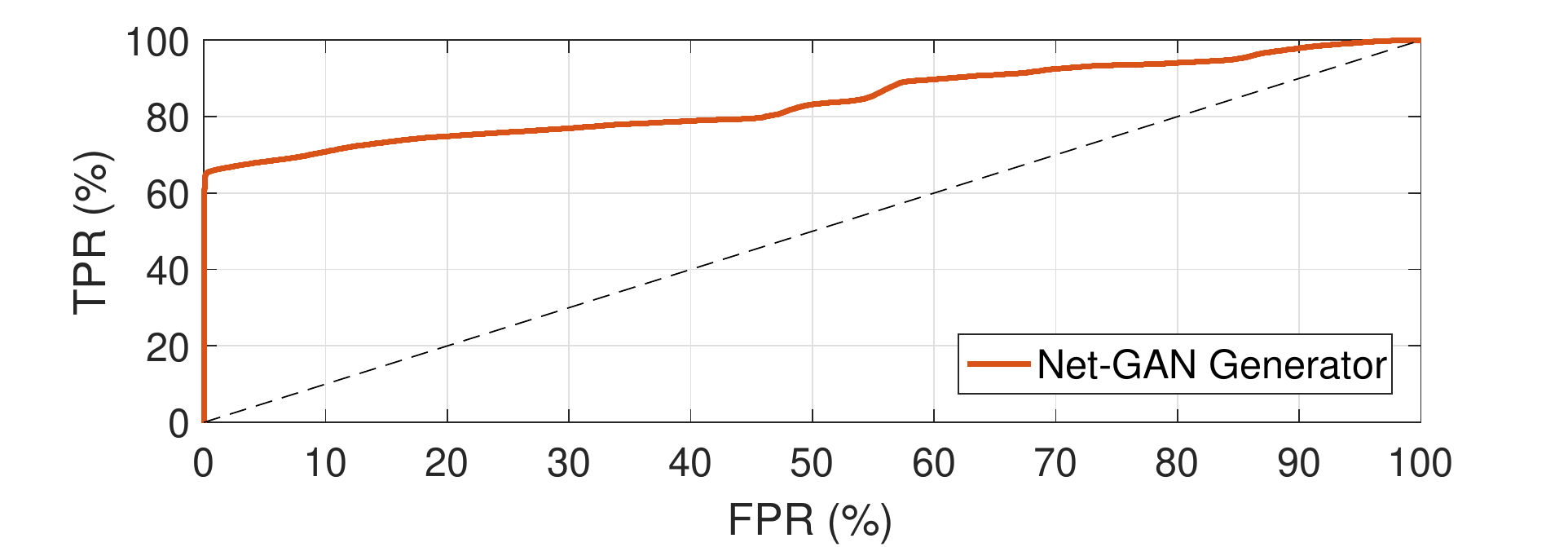}\\\vspace{-0.2mm}
\text{\scriptsize{(b) Detection performance with Net-GAN-G.}}\\
\includegraphics[width=0.95\linewidth]{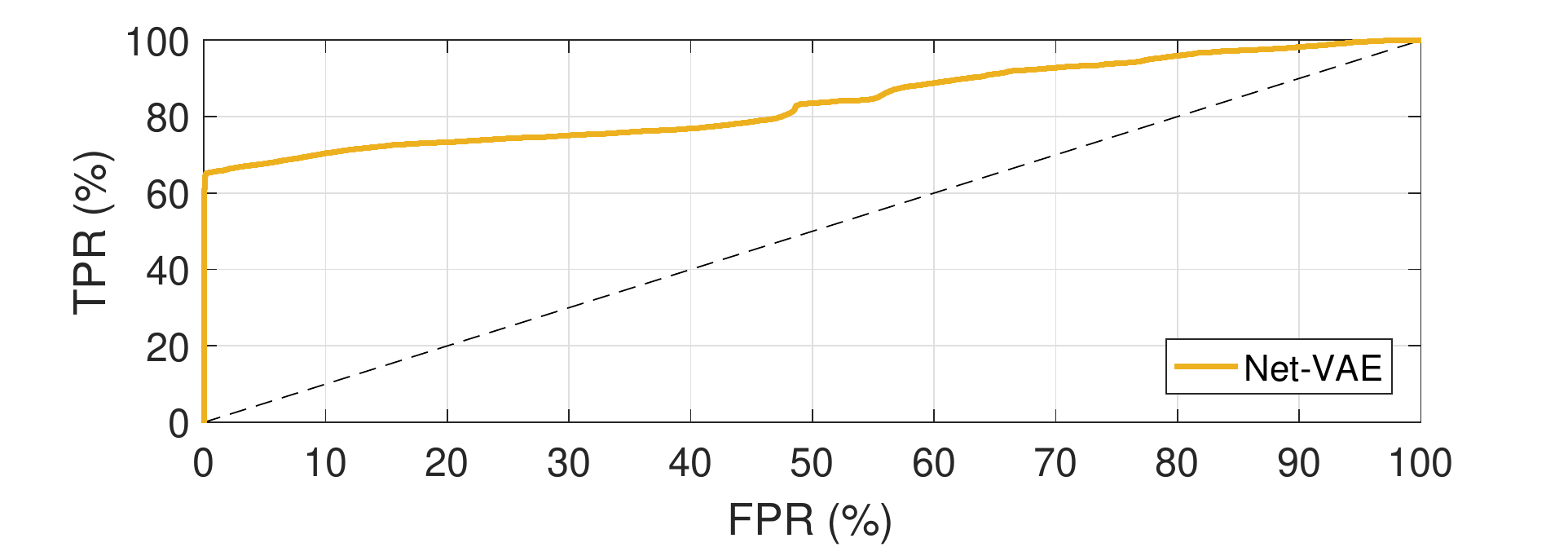}\\\vspace{-0.2mm}
\text{\scriptsize{(c) Detection performance with Net-VAE.}}\\
\end{array}$
\caption{Detection of anomalies in CPS data.}\label{ROC}
\end{figure}

In the case of VAE, the architecture is composed of the standard encoder and decoder functions which form the auto-encoder. An auto-encoder is a type of neural network used to learn both efficient representations of the input data, typically for dimensionality reduction -- the \emph{Encoder}, along with (re)generation models, which generate representations from the reduced encoding latent space as close as possible to its original input -- the \emph{Decoder}. Fig.~\ref{Net-VAE} depicts the Net-VAE architecture, which is composed of two data alignment and reconstruction layers -- to pre-process the time-series and post-process the auto-encoded samples, and two three-layer, feed-forward neural networks, representing the encoder and the decoder, respectively. The detection with Net-VAE is simply done through the \emph{residual loss} obtained by applying the trained VAE to the input testing sample $x$; if the difference between the input $x$ and its reconstruction $\hat{x}$ is greater than a detection threshold, an anomaly is declared. 

In terms of time-series preparation and processing, both Net-GAN and Net-VAE operate through a sliding window of $T$ samples -- using a unitary step. At each new step of the analysis, a matrix consisting of $n$ chunks of $T$ consecutive samples each is fed to the models, see Fig.~\ref{Net-VAE}. Finally, distance among time-series chunks is computed at a per-sample basis, and an anomaly is declared as soon as one or more of the samples deviate from the baseline by more than a detection threshold. To avoid false alarm due to spurious variations in the time-series, each sample $T$ generally represents a temporal aggregation of measurements, e.g., a moving average.    

\section{Preliminary Evaluation Results}\label{SIV}

We evaluate Net-GAN's detection performance on two different publicly available datasets, here referred to as CPS \cite{li2018anomaly} and SYN-NET \cite{sharafaldin2018}. The CPS dataset consists of synthetically generated attacks targeting industrial control systems, in particular a safe water treatment plant. It includes IoT sensor measurements for 51 different physical properties related to the plant and the water treatment process. In total, 946.722 samples are collected with a 1-second resolution, over 11 days. The SYN-NET dataset is a synthetically generated dataset for network intrusion detection, including normal operation traffic generated by a group of 25 users (e.g., HTTP/HTTPS browsing, FTP file transfer, SSH and mail, etc.), with controlled attacks over-imposed, of very different nature. In particular, we test Net-GAN for the detection of botnet traffic (0.2\% of total flows), DDoS attacks (4.3\%), port scan activity (16.6\%), and infiltration activity (only 36 flows). SYN-NET consists of more than a million flows -- 83\%/17\% benign/malign traffic. For the sake of completeness and performance-benchmarking, we also evaluate Net-VAE on the CPS dataset.

To show the generation capabilities of Net-GAN, Fig.~\ref{synthetic_ts} depicts some of the (min-max normalized) time series generated by the trained generator $G^*$, along with the corresponding real time-series. To reduce noise, samples are aggregated in time-windows of 10'. Time-series of higher magnitude are better reconstructed (TS$_1$ and TS$_2$), whereas noisy ones -- such as TS$_3$, are more difficult to track. 

Fig.~\ref{ROC} reports the detection performance achieved by Net-GAN and Net-VAE in the CPS dataset, in the form of ROC curves. Fig.~\ref{ROC}(a) reports the obtained results when using Net-GAN's discriminator $D^*$ as detector (Net-GAN-D), Fig.~\ref{ROC}(b) uses Net-GAN's generator $G^*$ as detector, and Fig.~\ref{ROC}(c) uses Net-VAE.

While results are preliminary and depend on the size and quality of the analyzed datasets -- we are currently working with bigger network measurement datasets, Net-GAN-D detects 56\% of the attacks with a FPR below 1\%, whereas both Net-GAN-G and Net-VAE detect close to 70\% of the attacks without false alarms. This shows that the generative capabilities of both approaches are extremely useful when it comes to detecting deviations. Still, the overall performance is rather poor for this scenario, which we believe is linked to the quality of the data. As reference, detection performance obtained in previous work \cite{li2018anomaly} for the same dataset is aligned with our results. 

\begin{figure}[t!]
\renewcommand{\arraystretch}{0.7}
\centering
$\begin{array}{cc}
\hspace{-4mm}\includegraphics[width=0.54\columnwidth]{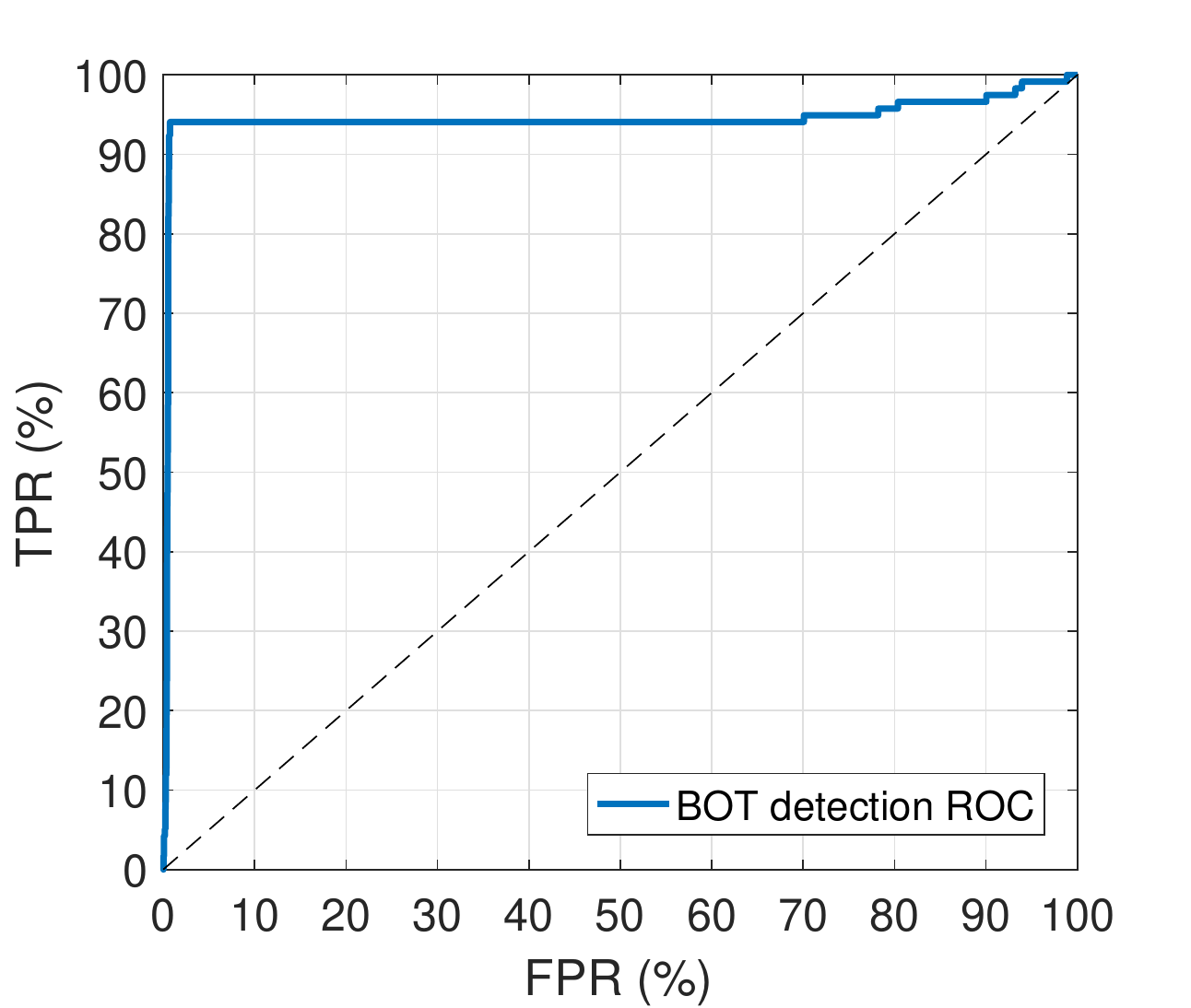} & \hspace{-4mm}\includegraphics[width=0.54\columnwidth]{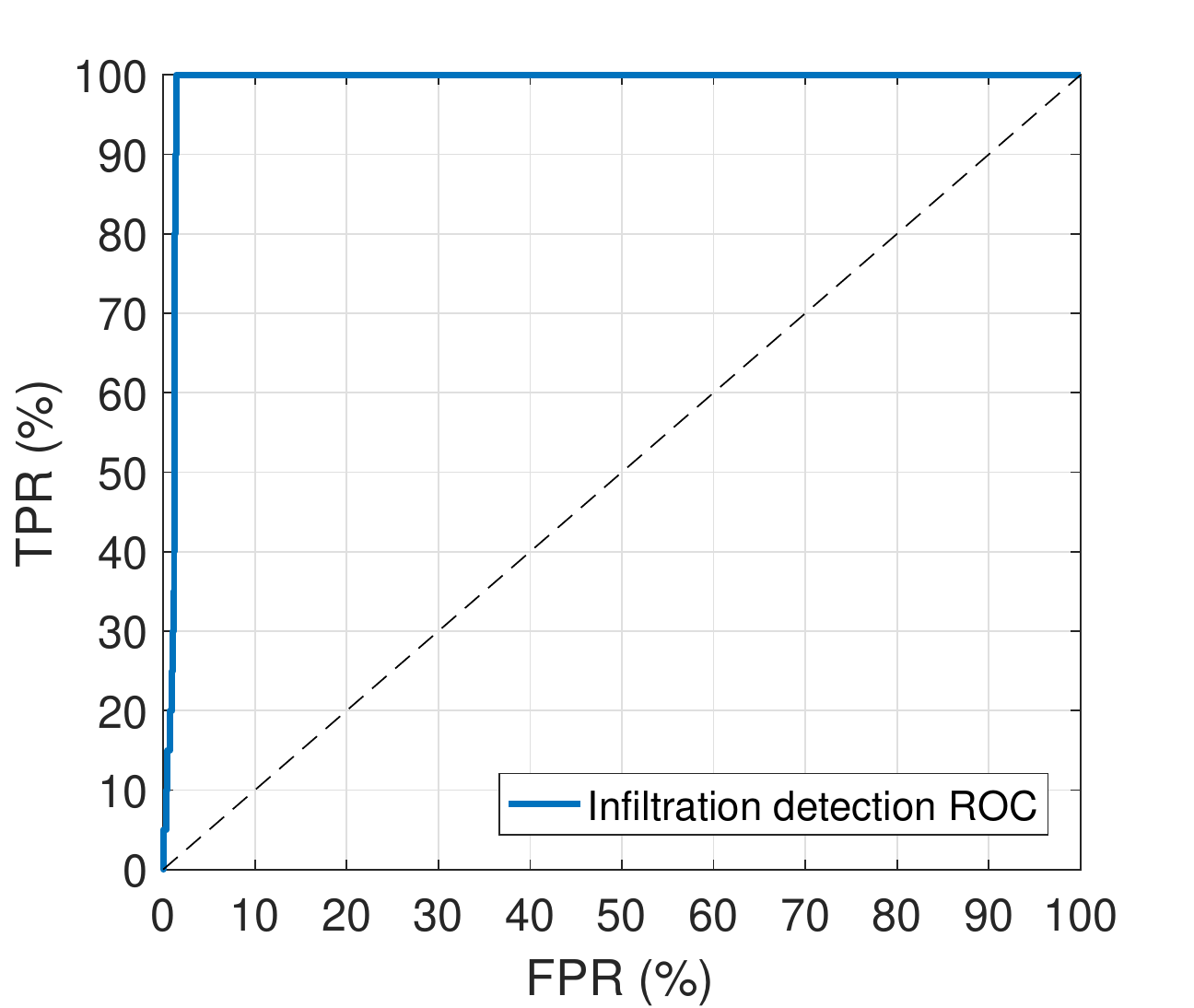}\\\vspace{-0.2mm}
\hspace{-4mm}\text{\scriptsize{(a) Botnet}} & \hspace{-4mm}\text{\scriptsize{(b) Infiltration}}\\\vspace{-0.2mm}
\hspace{-4mm}\includegraphics[width=0.54\columnwidth]{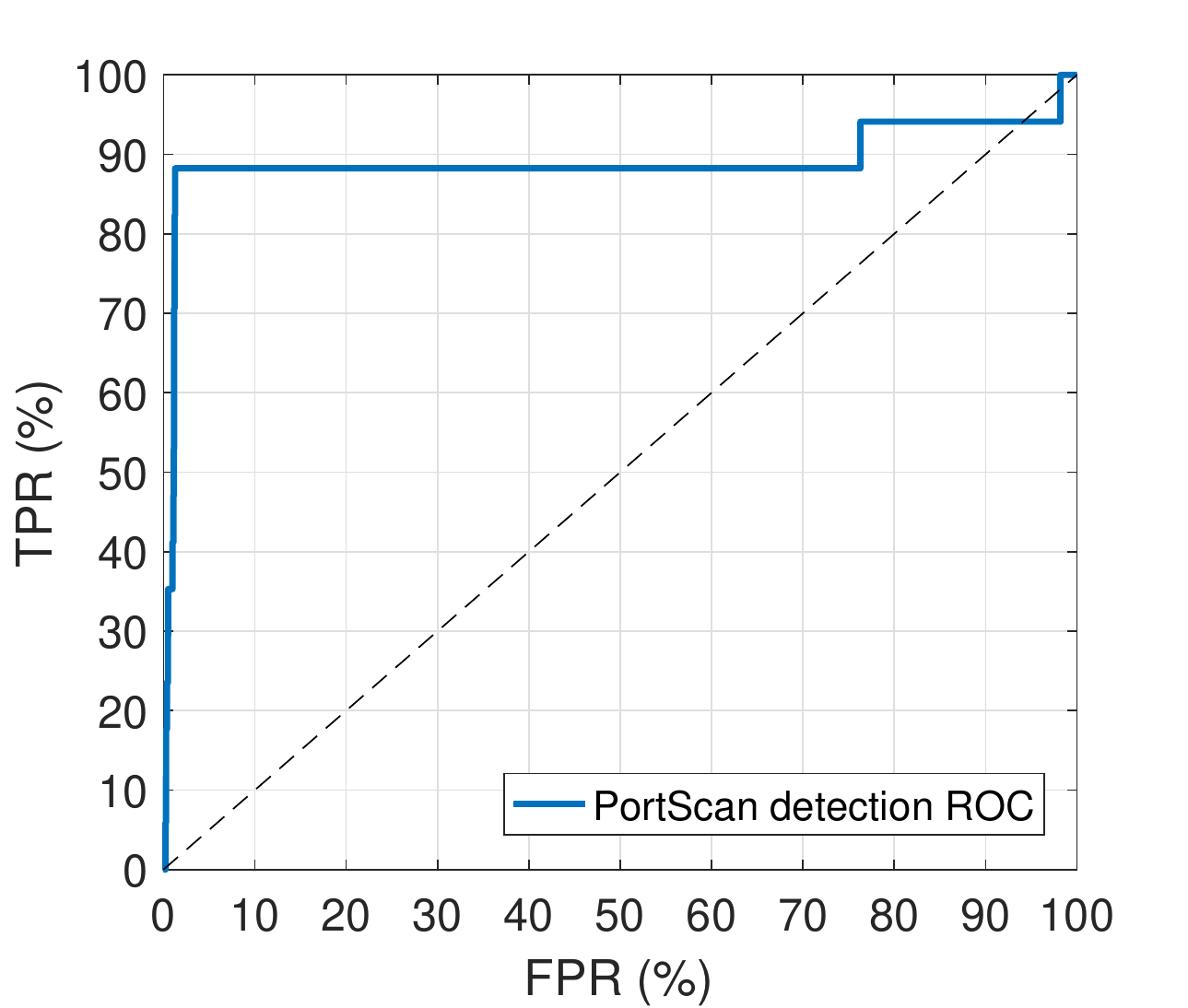} & \hspace{-4mm}\includegraphics[width=0.54\columnwidth]{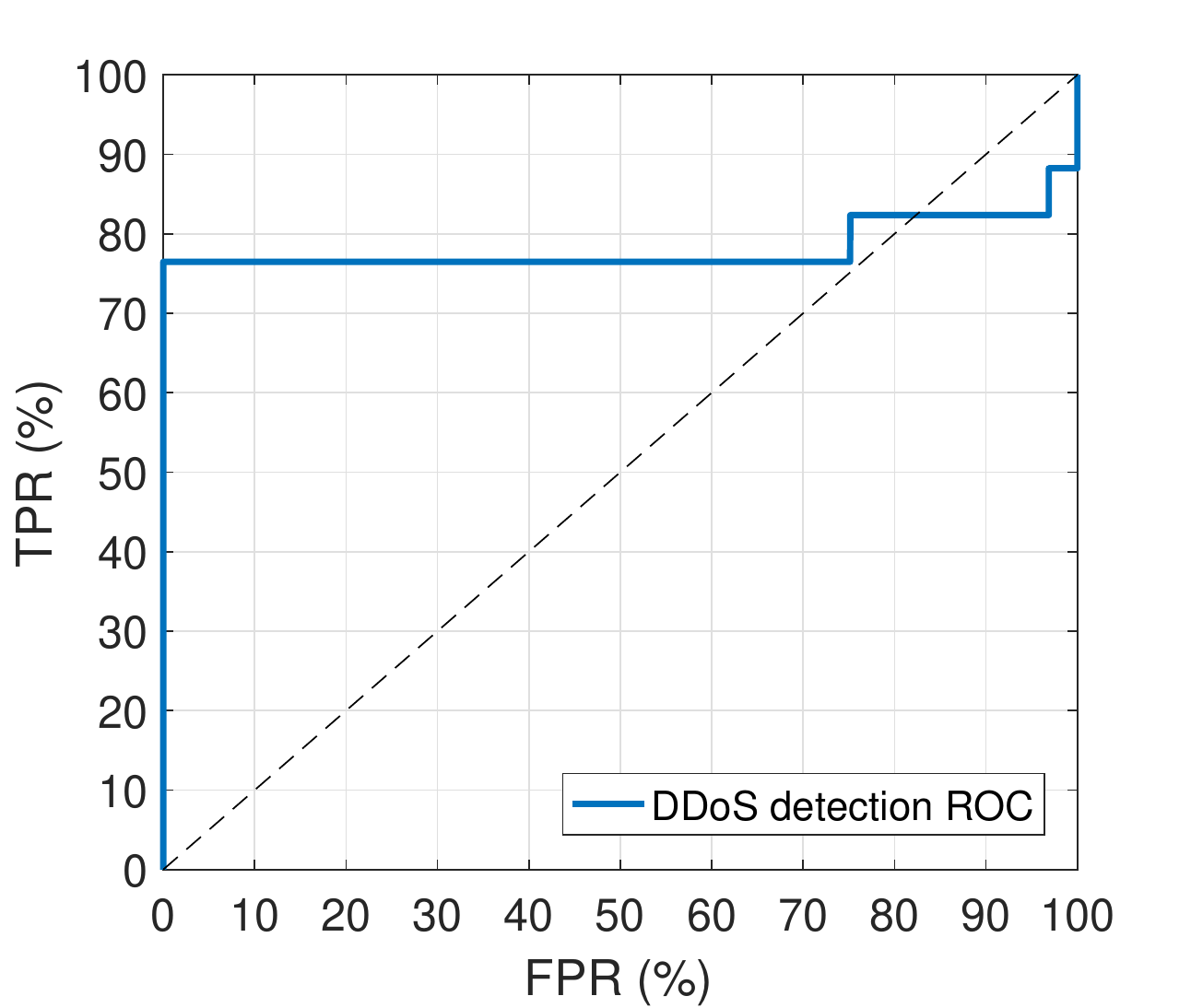}\\\vspace{-0.2mm}
\hspace{-4mm}\text{\scriptsize{(c) Port Scan}} & \hspace{-4mm}\text{\scriptsize{(d) DDoS}}\\\vspace{-0.2mm}
\end{array}$
\caption{Detection of attacks in SYN-NET data, using Net-GAN-D as underlying detection model.}
\label{roc_network_data}
\end{figure}

To conclude the paper, and to showcase the performance of Net-GAN in a different dataset, Fig.~\ref{roc_network_data} reports the detection performance of Net-GAN-D in the SYN-NET dataset, for the four considered network attacks. About 93\%, 100\%, 89\%, and 78\% of the attacks are detected with a FPR below 1\%, for botnet, infiltration, port scan, and DDoS, respectively, showing promising results in this specific scenario.

\balance

\section{Concluding Remarks}\label{SV}

In this paper, we have explored the application of modern generative models to the detection of anomalies in multivariate time-series. We have presented and evaluated \emph{Net-GAN}, a novel approach to network anomaly detection in time-series, using RNNs and GANs. Different from the state of the art, which traditionally focuses on univariate measurements, Net-GAN detects anomalies in multivariate time-series, exploiting temporal dependencies through RNNs. Net-GAN discovers the underlying distribution of the baseline, multivariate data, without making any assumptions on its nature, offering a powerful approach to detect anomalies in complex, difficult to model network monitoring data. We complemented Net-GAN with an alternative approach based on variational auto-encoders, which also represent a powerful and appealing model for the specific task.

The evaluation of both detection approaches in two different monitoring scenarios, including anomaly detection in IoT sensor data, and intrusion detection in network measurements, shows promising results. Besides the specific performance attained in terms of detection properties and generation of false alarms, generative-based models for anomaly detection might prove extremely useful when confronted with the monitoring of complex and highly-dimensional systems -- such as modern networks, where traditional modeling approaches would fall short to capture the underlying (joint) distributions of the data. A deeper and more comprehensive evaluation of Net-GAN and Net-VAE in real networking data at large scale is part of our ongoing work.

\section*{Acknowledgments}

This work has been partially supported by the ANII-FMV project with reference FMV\_1\_2019\_1\_155850 ``Anomaly Detection with Continual and Streaming Machine Learning on Big Data Telecommunications Networks'', and by Telef\'onica. Gast\'on Garc\'ia was supported by the ANII scholarship No. POS\_FMV\_2020\_1\_1009239.

\end{document}